\documentclass{llncs}
\usepackage{graphicx}
\usepackage{subfig}
\usepackage{makeidx}  
\begin{document}
\title{Prefrontal Cortex Motivated Cognitive Architecture for Multiple Robots}
\titlerunning{CIBRA}  
%
\author{Amit Kumar Mishra\inst{1} \and Abhishek Kumar\inst{2}
 \and Dipankar Deb\inst{3}  }
\authorrunning{Mishra et al.} 
%
\tocauthor{Amit Mishra, Abhishek Kumar, and Dipankar Deb}
\institute{University of Cape Town, South Africa,
\email{akmishra@ieee.org},
\and
Indian Institute of Technology Guwahati, India,
\and
GE Global Research, India.
}

\maketitle              

\begin{abstract}
In this paper, we  introduce a cerebral cortex inspired  architecture for robots in which we have mapped hierarchical cortical representation of human brain to logic flow and decision making process. 
Our work focuses on the two major features of human cognitive process, viz. the perception-action cycle and its hierarchical organization, and the decision making process. To prove the effectiveness of our proposed method, we  incorporated this architecture in our robot which we named as Cognitive Insect Robot inspired by Brain Architecture (CIRBA). 
We have extended our research to the implementation of this cognitive architecture of CIRBA in multiple robots and have analyzed the level of cognition attained by them.
\keywords{Formation, cognitive robotics}
\end{abstract}
\section{Introduction}
The field of robotics has seen tremendous development in past two decades. 
In the contemporary world robots are 
many times used in scenarios where discretion is  expected of them. This has brought a shift in paradigm from autonomous to evolutionary robotics and finally from evolutionary to cognitive robotics.

Harvey in 1992 talked on the issues in evolutionary robotics and advocated the use of artificial neural networks as the evolutionary architecture for robots \cite{Harvey92}. Few articles were also published, which talked about natural intelligence and human intelligence and attempts were made to explain behavior based artificial intelligence by building robots \cite{Christaller98}. 
As a result, working on the brain inspired architecture for robots became a necessity. Working in this direction, Soar cognitive architecture was used to build Adaptive Dynamics and Adaptive Perception for Thoughts (ADAPT) and implemented in a Pioneer mobile robot \cite{Benjamin04}. In another research, robot's task script was integrated with EM-ONE cognitive architecture \cite{Jung07}. 

Most of the robots built are task and environment specific. 
Secondly, most current works  are neither sufficient in bringing down the requirement of vast memory nor in reducing the complexity of algorithms. 
In the current work we propose the incorporation of the hierarchical organization of memory and knowledge and also the decision making method  of \emph{Homo sapiens} in robots. This will not only enable robots to adapt to any unknown environment but also make their behavior similar to human. 
Human cerebral cortex is the part of human brain responsible for the natural intelligence \cite{Christaller98} which we display and has always been a subject of study for researchers working in the field of cognitive science. In order to empower robots with human level intelligence, for the first time the idea of perception-action cycle of human beings \cite{Fuster04} was used by mapping different cortical regions of human brain to robots brain architecture \cite{Abhishek11}. In the current work we advocate the use of rational decision making process of human brain \cite{Strube98} in not just one robot but in a group of robots and study their behavior.


\begin{figure*}[!t]
\centerline{\subfloat[\footnotesize Levels of abstraction]{\includegraphics[width=2in]{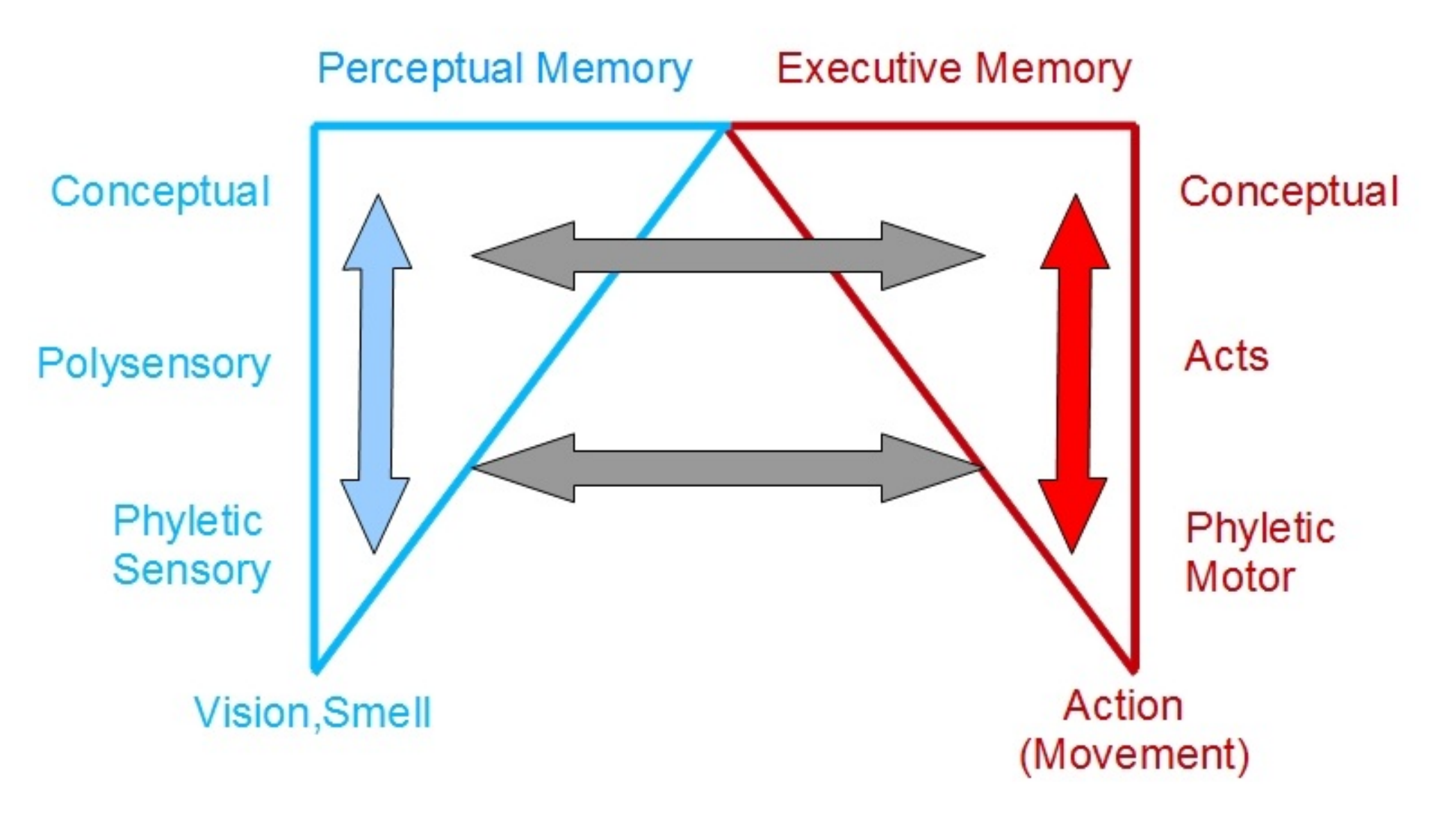} \label{fig1}}
\hfil
\subfloat[\footnotesize Perception-Action Cycle]{\includegraphics[width=2in]{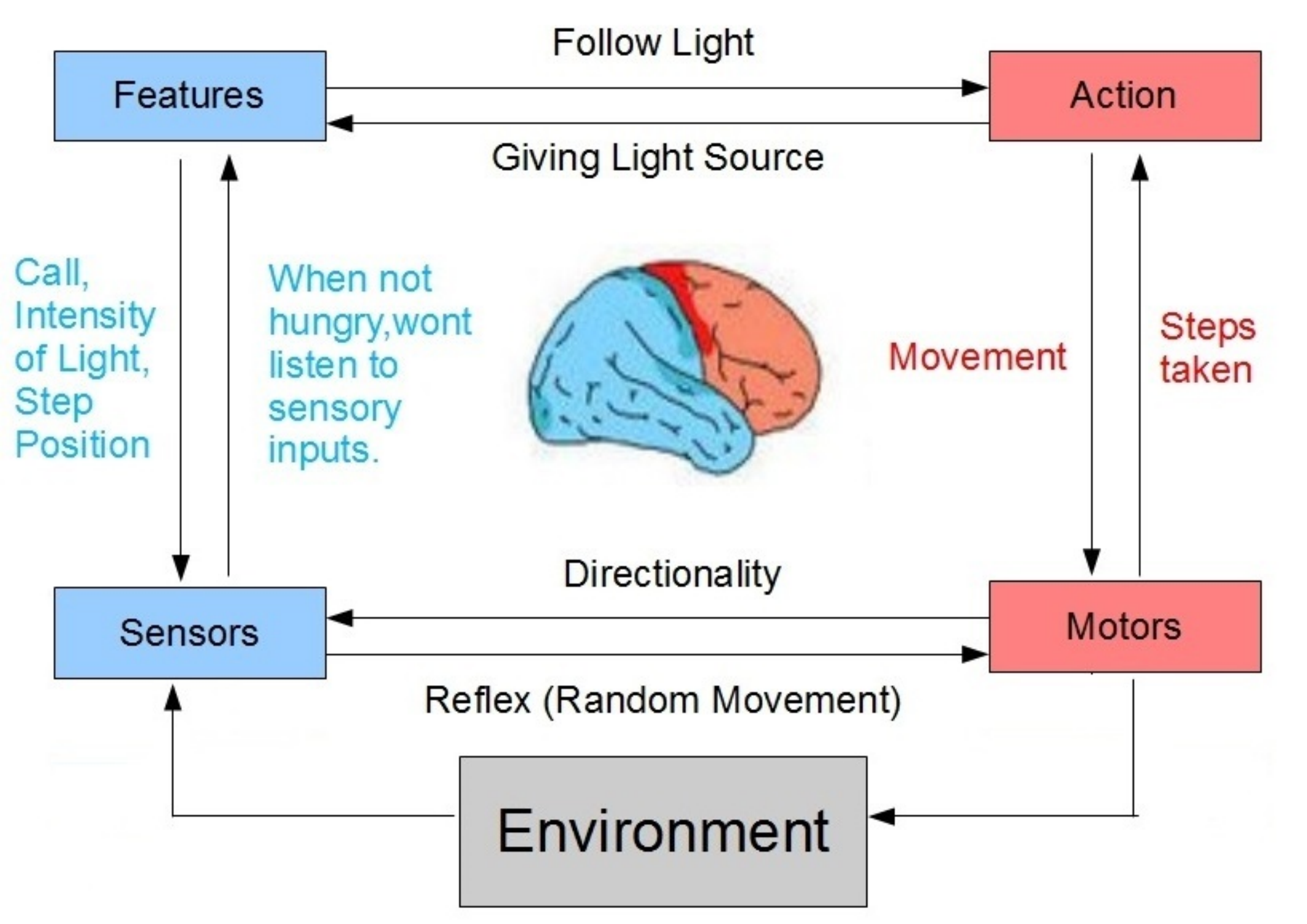} \label{fig2}}}
\caption{Cognitive Architecture used in CIRBA \cite{Abhishek11}}
\label{fig_1}
\end{figure*}

\section{Overview of Previous Works}

Cognitive robotics is a fusion of robotics, evolutionary modeling and psychology. It is endowing robots with the power to reason, act, react and hence, adapt to unknown and changing environment. 
In most of the current works in this field there remains a common problem, i.e. the robots designed perform only specific task and only in the given environment. In one of our previous works \cite{Abhishek11} we attempted  to make robots behavior independent of their surrounding, used perception-action cycle of human beings in them. There are two major aspects of human brain architecture.

\begin{itemize}
 \item Perception-action cycle and
 \item Rational decision making.
\end{itemize}

\subsection{Cognitive Insect Robot inspired by Brain Architecture}

According to Fuster, perception-action cycle is the circular flow of information from environment to sensory structures, to motor structures, back again to the environment, then to sensory structures and so on while manifesting a goal oriented behavior \cite{Fuster04}. During the time of birth, very little information is present in a new born. This memory is actually phyletic memory acquired by humans through evolution and is close to the sensory area. By experience a concept is developed in the higher hierarchical level and accordingly actions are performed with feedback at every level of abstraction. At each level of abstraction, the processed output depends on:
\begin{itemize}
 \item Information derived from sensory signals and
 \item Processing of global aspects of the result at that level in upper frontal areas \cite{Fuster04}.
\end{itemize}
Wherever complex information is to be processed, conceptual layer of both perceptual and executive memory comes into picture, whereas phyletic memory handles any task whenever sensory data requiring instant response are received.

This concept was successfully incorporated in CIRBA (Fig. \ref{fig_1}). One of the remarkable outcomes of using human cognitive architecture was, CIRBA gaining the ability to perform goal oriented task with minimum information fed in its phyletic memory (its level of happiness). In the experiment explained in \cite{Abhishek11} through evolution CIRBA learnt that light is its food and defined searching light (its food) as its goal. Through experience it learns whether responding to call made from its home is beneficial for it or not.

From a void conceptual memory, CIRBA during its life span learns and unlearns information required for decision making through experience. As it can be seen from Fig. \ref{fig2} inputs are received from environment through sensors and data (intensity of light received or call from home) is sent to the higher cortical level in the perceptual wing but if CIRBA is not hungry then it starts neglecting these sensory data. On the topmost level of hierarchy after the features are decided, decision is made, whether to follow light or not. The action block then decides the movement of the motors, in the meantime, information regarding the steps taken is continuously sent to it to monitor the movement of the robot. In the lowest level of perception-action cycle, motors provide directionality to the sensors and if the robot gets stuck somewhere then the sensors call interrupt and without the intervention of higher level motors, perform random motion to come out from the undesirable location.

\subsection{Rational Decision Making in Human beings}

The seeds of learning human behavior governed by rational decision making were sown in early forties. But still the rational decision theory has not changed and it says that people make choices so as to maximize their profits. In psychology, each behavioral alternative is assigned a utility, or (subjective) valence, V \cite{Strube98}. According to Lewin, Dembo, Festinger and Sears, the sum of the valences of the possible outcomes of an action with each outcome weighted by its estimated probability of occurrence can be calculated from:
\begin{equation}\label{eq1}
    V_r = V_s \times P_s + V_f \times P_f
\end{equation}
where, V$_r$ is the resulting valence of an action, split into valence and probability of success (index s) and failure (index f). The probabilities $P_s$ and $P_f$ add up to 1; the valence for failure is usually negative. The theory says that the alternative available with maximum $V_r$ is chosen by any person \cite{Lewin44}.

Atkinson brought a slight change in valence of action and presented it as the product of personal motive strength M and incentive strength I. The well-known model by Atkinson and Feather \cite{Atkinson66} integrated this view with Lewin et al.s formula to define the resultant tendency T as:
\begin{equation}\label{eq2}
    T_r = M_s \times P_s \times I_s + M_f \times P_f \times I_f
\end{equation}
Here also the alternative with maximum $T_r$ is chosen by any person.
\section{Decision Making and dynamics of CIRBA}

Keeping the functionality of CIRBA same, we propose the use of rational decision making of human beings in CIRBA. During its time of evolution with nothing in its innate memory, its motion is random. But each time, after receiving light beyond certain threshold (happiness level) its probability to search light increases as it can be seen from \ref{eq3}.

\begin{equation}\label{eq3}
    P_{sl} = M + P_{exp} \times \Delta{L}
\end{equation}
where,$P_{sl}$ is the probability to search light, M (mood of robot) is the random number generated by robot between 0-1, $P_{exp}$ (previous experience) is value stored in CIRBA's memory from past experience, i.e. whether it had experienced satisfaction (positive value) or not (negative value) after performing the action and $\Delta{L}$ (Learning parameter) is the resolution deciding how fast we want robot to learn.

After repeated learning process $P_{sl}$ becomes 1 and it is after this point of time, CIRBA's motion becomes goal oriented instead of random. Now its basic aim of life is to look for light. If a call is made from its base or origin then it has to decide whether to respond to the call or keep looking for light.This decision is made using \ref{eq4}.

\begin{equation}\label{eq4}
    P_{rsp} = M + P_{exp} \times \Delta{L}
\end{equation}
where, $P_{rsp}$ : Probability to respond. Here it is to be noted that we have taken into account the mood factor as it was proposed by Atkinson \cite{Atkinson66}. This mood is nothing but a random number generated through hardware noise. Whenever $P_{rsp}$ is greater than 0.5, CIRBA gives a positive response to the call, i.e. returns to its home. Each time it is fed with light after giving positive response, $P_{exp}$ increases otherwise decreases.

Dynamics of CIRBA can be modeled using the equation of spring damper:
\begin{equation}\label{eq5}
    F = m\dot{\omega} + k \theta + c \omega
\end{equation}
And as pd controller:
\begin{equation}\label{eq6}
    F = k_p(\theta_{ref}-\theta)+k_d\omega
\end{equation}
\begin{equation}\label{eq7}
    \dot{\theta} = \omega
\end{equation}
$\theta_{ref}$ is the desired direction for robot.
\\Using \ref{eq5} and \ref{eq6} and eliminating F, we get,
\begin{equation}\label{eq8a}
    \dot{\omega} = [-(k_d+c)\omega-(k_p+k)\theta+k_p\theta_{ref}]/m
\end{equation}
\begin{equation}\label{eq8b}
    \dot{\theta} = \omega
\end{equation}
Equations \ref{eq8a} and \ref{eq8b} are the state equation of CIRBA.

While looking for light it also keeps track of obstacles it has encountered with and stores their position. $k_p$ and $k_d$ is also variable in case of CIRBA.
\begin{equation}\label{eq9a}
    k_p=k_{po}+\Delta{L_1}\times{n}
\end{equation}
\begin{equation}\label{eq9b}
    k_d=k_{do}+\Delta{L_2}\times{n}
\end{equation}
where, $k_{po}>k_{do}$, $k_{p}>(k+\epsilon_1)$ and $k_d>(k+\epsilon_2)$
\\$\epsilon_1$ and $\epsilon_2$ are determined experimentally and are greater than 1, so that $k_p$ and $k_d$ are able to rotate robot towards $\theta_{ref}$. Otherwise, $k$ brings the robot to $0$ radian and prevents it from turning. $\Delta{L_1}$ and $\Delta{L_2}$ are probabilistically determined. Random number is generated between (0,1) and each time the robot visits any place, its $k_p$ and $k_d$ increases until it reaches certain saturation value $k_{pmax}$ and $k_{dmax}$. This type of response of CIRBA is analogous to human behavior. 

\begin{figure*}[!t]
\centerline{\subfloat{\includegraphics[width=2.5in]{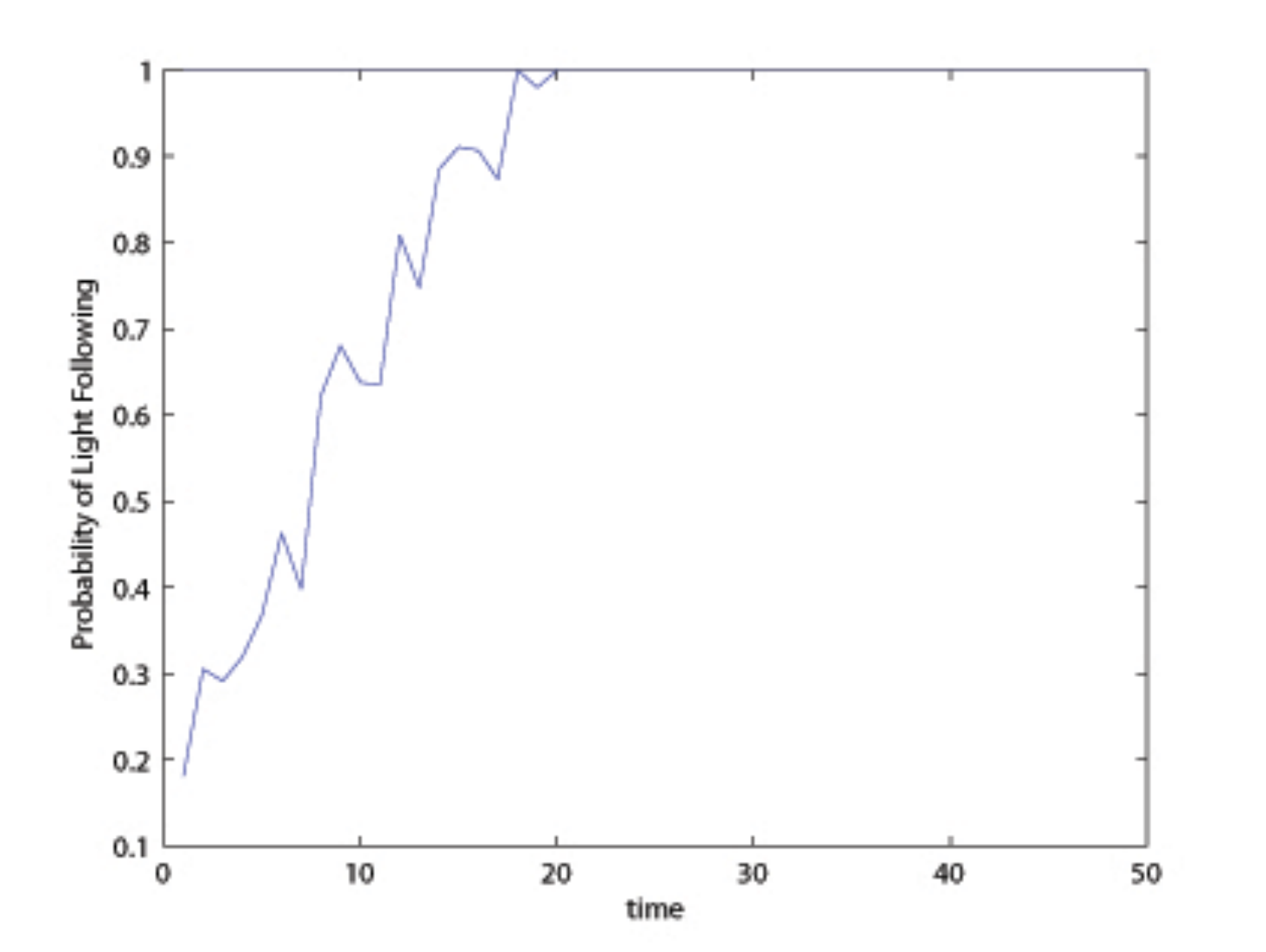} \label{fig}}
\hfil
\subfloat{\includegraphics[width=2.5in]{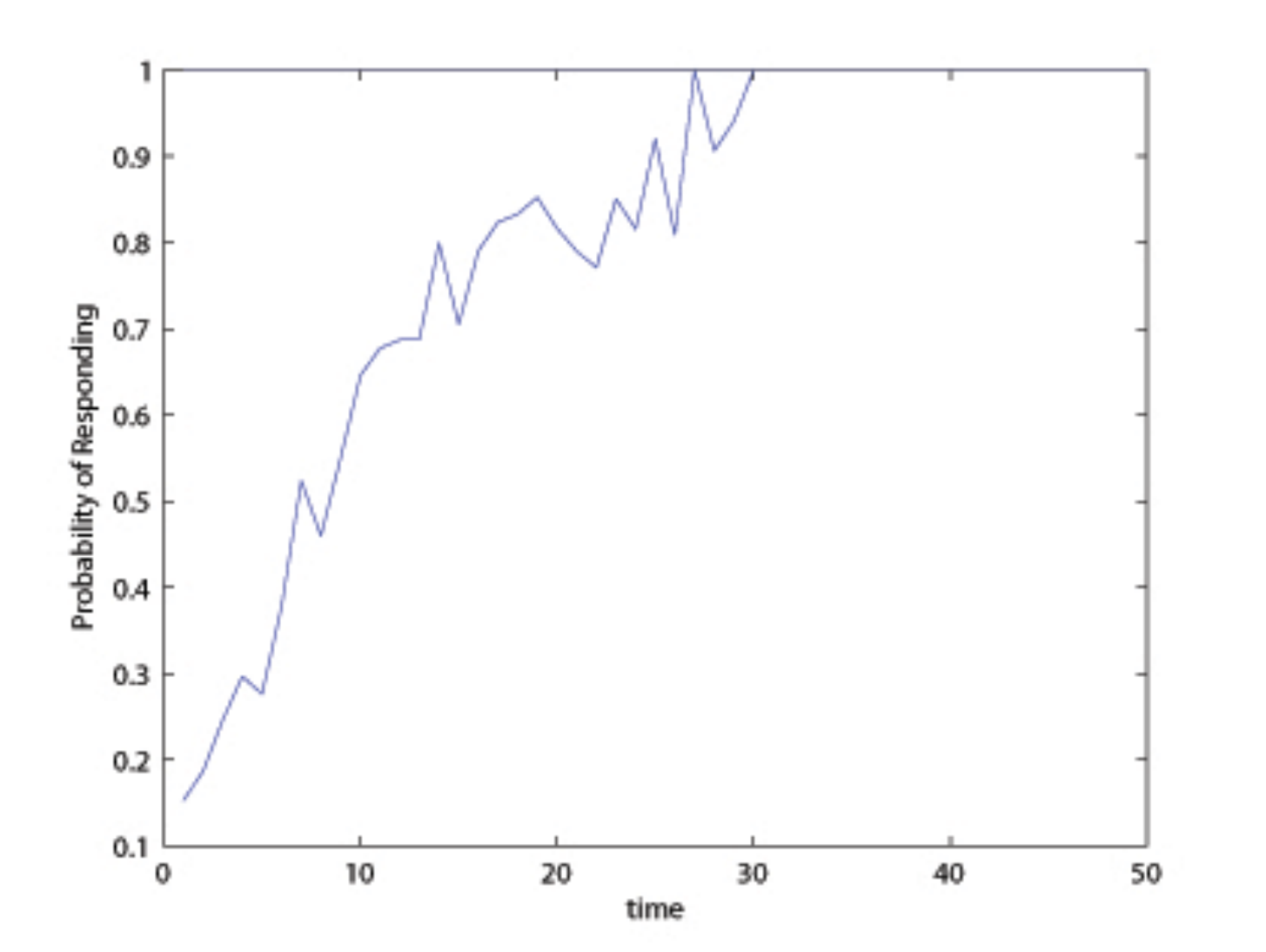} \label{fig}}}
\caption{Variation of probability of searching light according to (\ref{eq3}) and probability of responding to call according to (\ref{eq4})}
\label{Probability}
\end{figure*}

\section{Use of Cognitive Architecture in Multiple Robots}

The discussed cognitive architecture is further implemented in three robots. These robots collectively move in triangular formation with one of them becoming 'the Lead Robot'. Robot receiving the maximum intensity of light becomes the Lead Robot but the decisions are made collectively by all the three members; i.e. learning parameter and mood of all three robots affects the decision making.

\begin{equation}\label{eq13}
    P_{rsp} = \sum_{i=1}^{3}m_i/3 + \sum_{i=1}^{3}({p_{exp_i} \times \Delta{L}})/3
\end{equation}
where, $P_{rsp}$ is probability to respond, $m_i$ is mood of the $i^{th}$ robot and $p_{exp_i}$ is previous experience of the $i^{th}$ robot.

Follower robots align themselves in accordance with the Lead Robot. Change in the orientation of the leader is continuously monitored by the other two using proximity sensors. $\theta_{ref}$ for them is the instantaneous $\theta$ of the pivot robot. In order to get this $\theta$, proportional gain ($k_{pg}$) and derivative gain ($k_{dg}$) of the group is calculated using the following formula:
\begin{equation}\label{eq14}
    k_{pg}=\sum_{i=1}^{3}k_{pi}/3 ~~~~ k_{dg}=\sum_{i=1}^{3}k_{di}/3
\end{equation}

where, $k_{pi}$ is proportional gain of individual robot calculated from (\ref{eq9a}) and $k_{di}$ is derivative gain of individual robot calculated from (\ref{eq9b}). These two gains of individual robot is calculated by themselves, then Robot1 transmits its gain to Robot2 and Robot2 sends both the gains to Robot3 which is the pivot robot. Gain for the group is calculated by it and then broadcasted to both robots.

\begin{figure*}[!t]
\centerline{\subfloat{\includegraphics[width=2.5in]{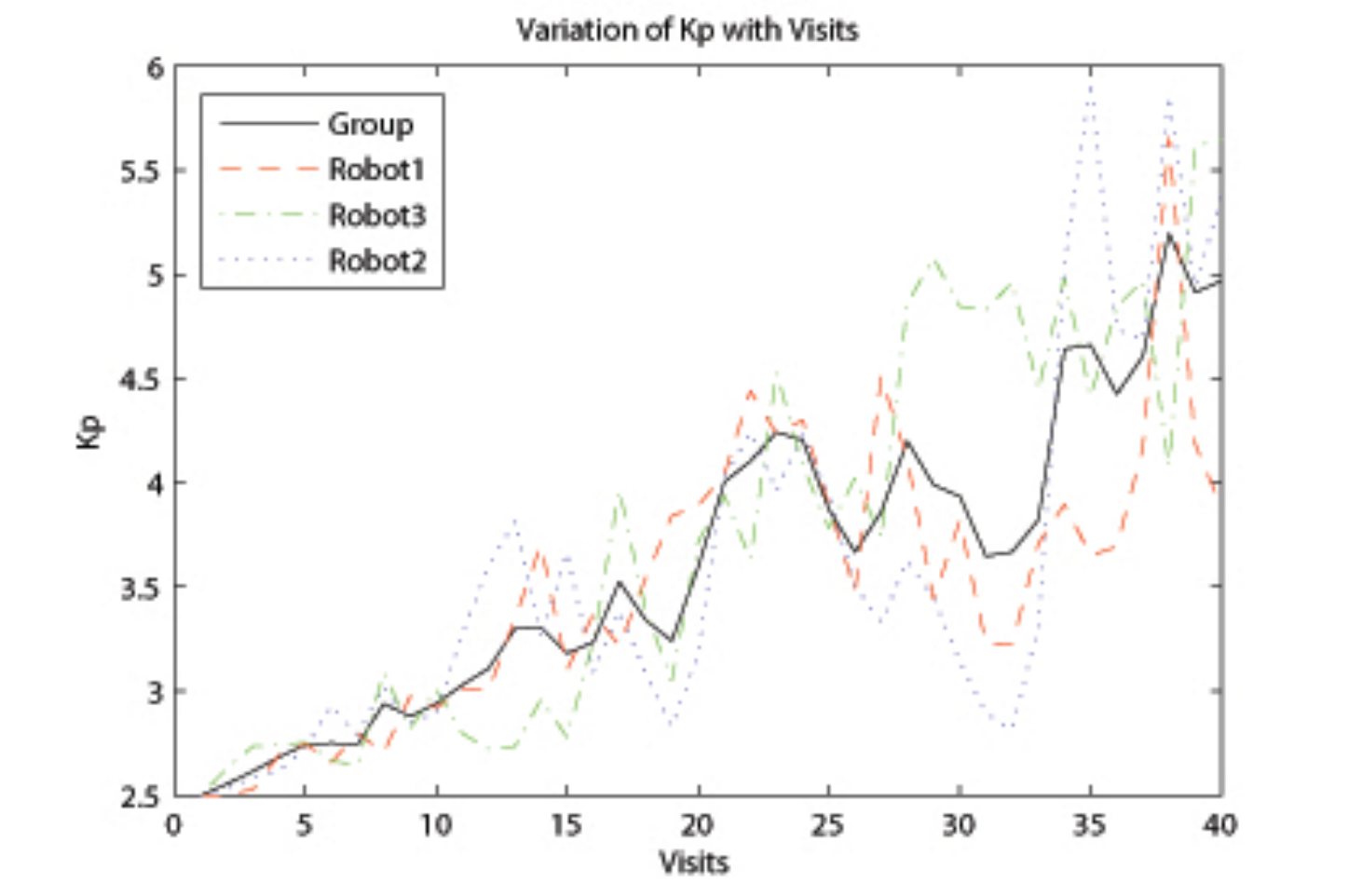} \label{k_p}}
\hfil
\subfloat{\includegraphics[width=2.5in]{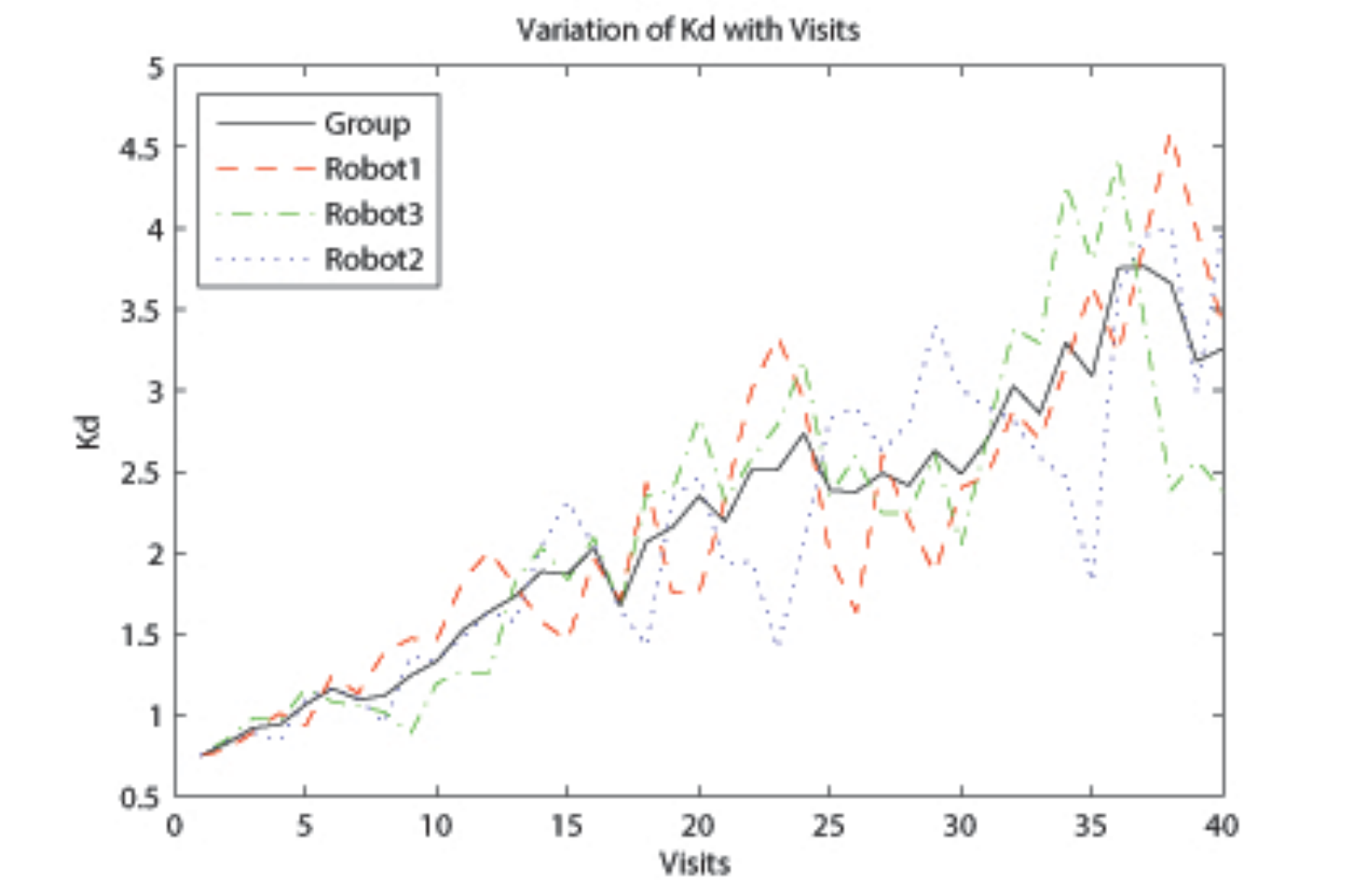} \label{k_d}}}
\caption{Variation of $k_p$ w.r.t number of visits, according to (\ref{eq9a}), (\ref{eq9b}) and (\ref{eq14})}
\label{fig}
\end{figure*}

\section{Results}

CIRBA's ability to learn is illustrated in Figure \ref{Probability}. 
t can be seen that initially probability of searching light or responding to call is very low but gradually through experience it learns about its food and also the reward it receives on giving positive response to the call, resulting in increase in probability; which eventually becomes one. Few dips in the graph can also be observed, the ones with larger magnitude represent unlearning of event when reward is not given to it, whereas the smaller dips show reluctance of CIRBA to responding to the call due to its mood.

If we try to analyze this behavior of CIRBA through perception-action cycle, we will see that this learnt information is stored in CIRBA's conceptual memory, but during its life time no change is brought in its phyletic memory. Even after gaining experience, during the time of evasive action (when it is stuck somewhere and reflex action is to be performed) it uses its reflexes to come out from that undesirable situation, i.e. only phyletic memory is used and conceptual memory is not used in decision making.

When working in group, values of $k_p$ and $k_d$ gets modified which can be easily deduced from Figure \ref{k_p} and Figure \ref{k_d}. Saturation level of $k_p$ and $k_d$ is experimentally determined. If their value is chosen less than $k$ in \ref{eq5}, the robot is not able to turn and as it happens with a spring, after few attempts it settles back to its original orientation. When values of $k_p$ and $k_d$ are chosen high, even for mild changes in the angle, turning becomes very haphazard. Values of these gains taken in our experiment are:
\\$k_{do} = 0.5 $, $k_{po} = 2.5 $, $k_{ds} = 5 $, $k_{ps} = 9 $ and $k = 0.1$
\\where, $k_{do}$ and $k_{po}$ represent initial values and $k_{ds}$ and $ k_{ps}$ define the maximum limit of $k_p$ and $k_d$.

From the graphs (\ref{k_p}) and (\ref{k_d}) it can be verified that $k_p$ and $k_d$ started from 2.5 and 0.5 respectively and after visiting the same place 40 times, these values increase to 7 and 4.5. When the robots visit the same place again and again, they become aware of the surrounding; therefore there is an increase in their $k_p$ and $k_d$ values. For individual robots, there is a lot of fluctuation in the increase of these proportional and derivative gains which arise due to variation in their mood. But when they work in group, it is observed that these fluctuations get averaged out.

\begin{figure}[t]
{\includegraphics[width=3.5in]{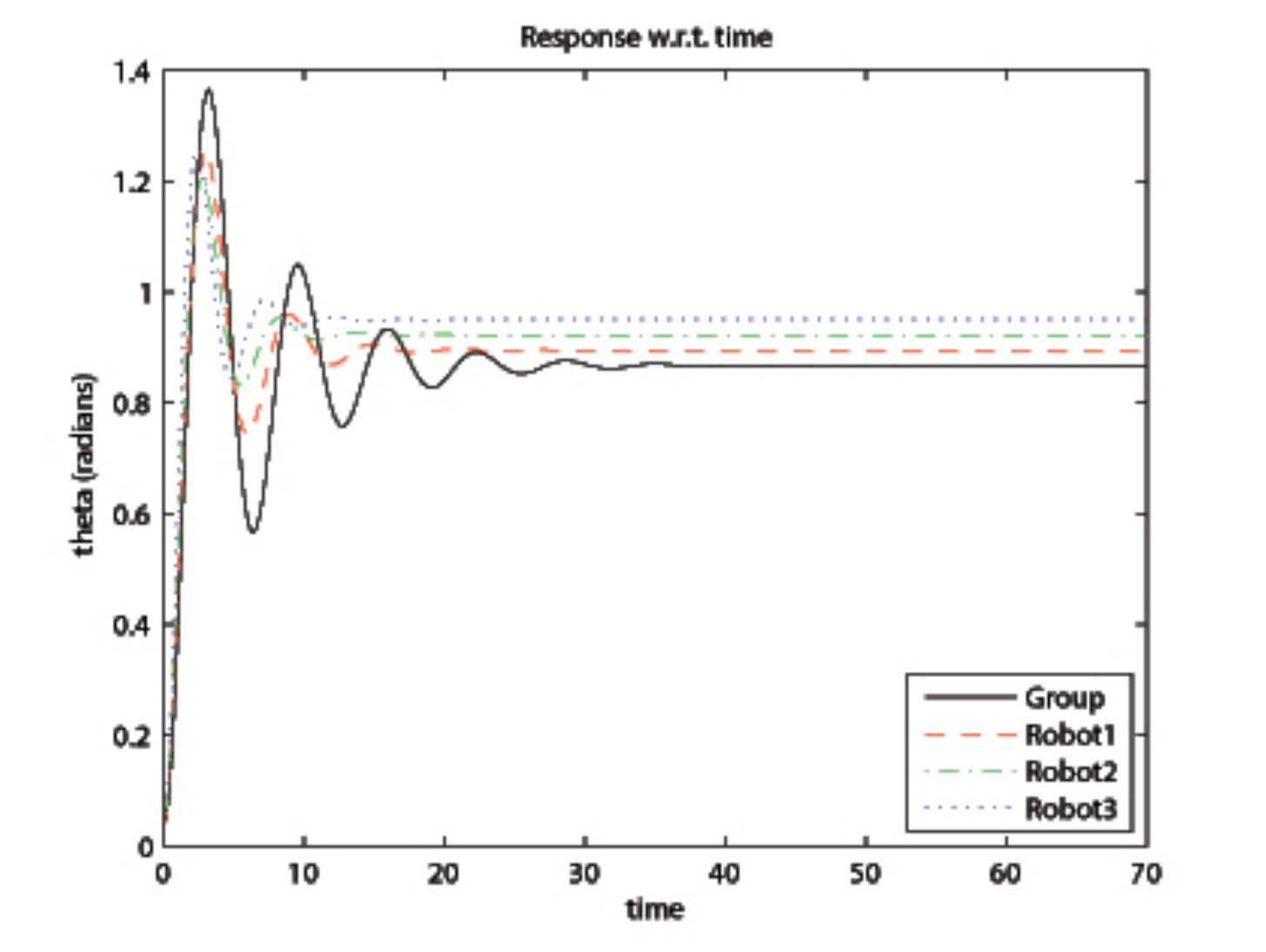} \label{fig}}
\caption{Response of CIRBA after different visits}
\label{theta}
\end{figure}

With the above values, CIRBA's turning ability at places which it has previously visited changes and the response of the Leader Robot as calculated from (\ref{eq8a}) and (\ref{eq8b}) is shown in Figure \ref{theta}. It can be observed that turning becomes more swift as we move from first -$>$ tenth -$>$ twentieth -$>$ thirtieth visit.
Here, initial value of $\theta$ is 0.04 radians and later on the robot reorients itself to $\theta_{ref}$ which is 0.9.

\section{Conclusion}

In most of the existing cognitive architecture, lots of information is already fed in the memory of any robot which cannot be exactly called as learning, whereas in our case we have tried to analyze CIRBA's response by feeding minimum amount of initial information in it. Apart from this, CIRBA's ability to show both concept based and reflex action is a remarkable achievement. In the cognitive scale CIRBA achieved a high score of 18.3 in ConsScale (www.consscale.com). Moreover, while using multiple robots, we shared their topmost hierarchical layer and tried to get benefit from them by making the robots work in a group and hence cover a wider area while searching for light. Moving ahead in this direction we are planning to build a platform capable of exhibiting human like behavior and hence, aid researchers in studying human nature.

\end{document}